\documentclass[letterpaper, 10 pt, conference]{ieeeconf}  

\IEEEoverridecommandlockouts                              

\usepackage{caption}
\usepackage[T1]{fontenc}
\usepackage{amsfonts}
\usepackage[cmex10]{amsmath}
\DeclareMathOperator*{\E}{\mathbb{E}}
\usepackage[hidelinks]{hyperref}
\usepackage{graphicx}
\usepackage{algorithm}%
\usepackage{algpseudocode}%
\title{\LARGE \bf
I am Robot: Neuromuscular Reinforcement Learning to Actuate Human Limbs through Functional Electrical Stimulation
}

\author{Nat Wannawas$^{1}$, Ali Shafti$^{1}$, and A. Aldo Faisal$^{1}$
\thanks{$^{1}$NW, AS, and AAF. are with the Brain and Behaviour Lab: Dept. of Bioengineering and Dept. of Computing, Imperial College London, SW7 2AZ, London, UK. Address for Correspondence {\tt\small aldo.faisal@imperial.ac.uk}}%
}

\begin{document}

\maketitle
\thispagestyle{empty}
\pagestyle{empty}

\begin{abstract}
Human movement disorders or paralysis lead to the loss of control of muscle activation and thus motor control. Functional Electrical Stimulation (FES) is an established and safe technique for contracting muscles by stimulating the skin above a muscle to induce its contraction. However, an open challenge remains on how to restore motor abilities to human limbs through FES, as the problem of controlling the stimulation is unclear. We are taking a robotics perspective on this problem, by developing robot learning algorithms that control the ultimate humanoid robot, the human body, through electrical muscle stimulation. Human muscles are not trivial to control as actuators due to their force production being non-stationary as a result of fatigue and other internal state changes, in contrast to robot actuators which are well-understood and stationary over broad operation ranges. We present our Deep Reinforcement Learning approach to the control of human muscles with FES, using a recurrent neural network for dynamic state representation, to overcome the unobserved elements of the behaviour of human muscles under external stimulation. We demonstrate our technique both in neuromuscular simulations but also experimentally on a human. Our results show that our controller can learn to manipulate human muscles, applying appropriate levels of stimulation to achieve the given tasks while compensating for advancing muscle fatigue which arises throughout the tasks. Additionally, our technique can learn quickly enough to be implemented in real-world human-in-the-loop settings.
\end{abstract}

\section{INTRODUCTION}
\label{sec:introduction}
Motor impairments and disabilities result in humans being unable to, or having limited ability in activating their muscles to move their own limbs, heavily impacting their independence and quality of life. Solutions have been conceived within robotics, in the form of assistive and rehabilitation robots \cite{Abidemi2018} as well as wearable and exoskeleton robots \cite{Shin2011, Wang2020}, to have robot actuators fulfil tasks for the users, or manipulate the human user's limbs for them \cite{Shafti2019}. While these approaches have shown success in restoring some of the user's capabilities through assistance, they do not make use of the potential for actuation within the user's own muscles. Functional Electrical Stimulation (FES) is a method to activate human muscles through external stimulation, leading to actuation of the connected limbs, and is used in a limited capacity in some assistive devices \cite{Dunkelberger2020}. Mastering neuromuscular control through FES would enable us to control a human body through its muscles in the same way we control robots through motors and actuators. This requires two open problems, coping with the non-stationarity of muscles and learning to control their non-linear responses so as to enable safe, robust multi-joint control. We focus here on the first challenge, as the second can be tackled, as we know from robotics, using Reinforcement Learning.

\begin{figure}[]
    \centering
    \includegraphics[width=0.9\columnwidth]{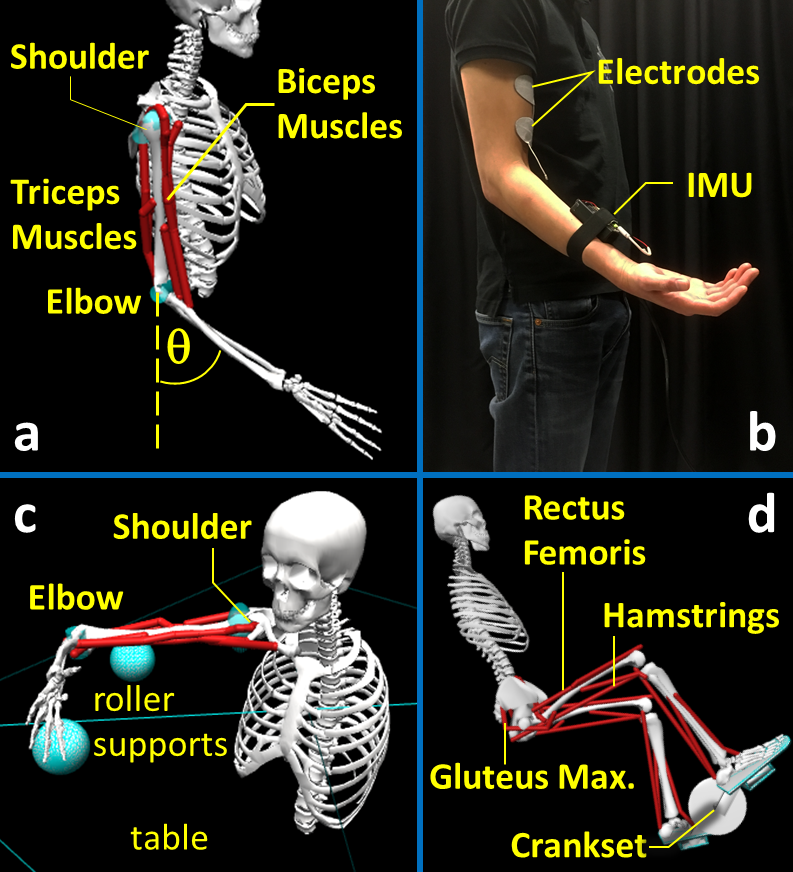}
    \caption{Our 3 scenarios for FES control:  (a) arm vertical motion in simulation (b) and human volunteers, (c) arm horizontal motion and (d) cycling in simulation.}
    \label{fig:setup}
\vspace{-5pt}
\end{figure}

\begin{figure}[]
    \centering
    \includegraphics[width=0.95\columnwidth]{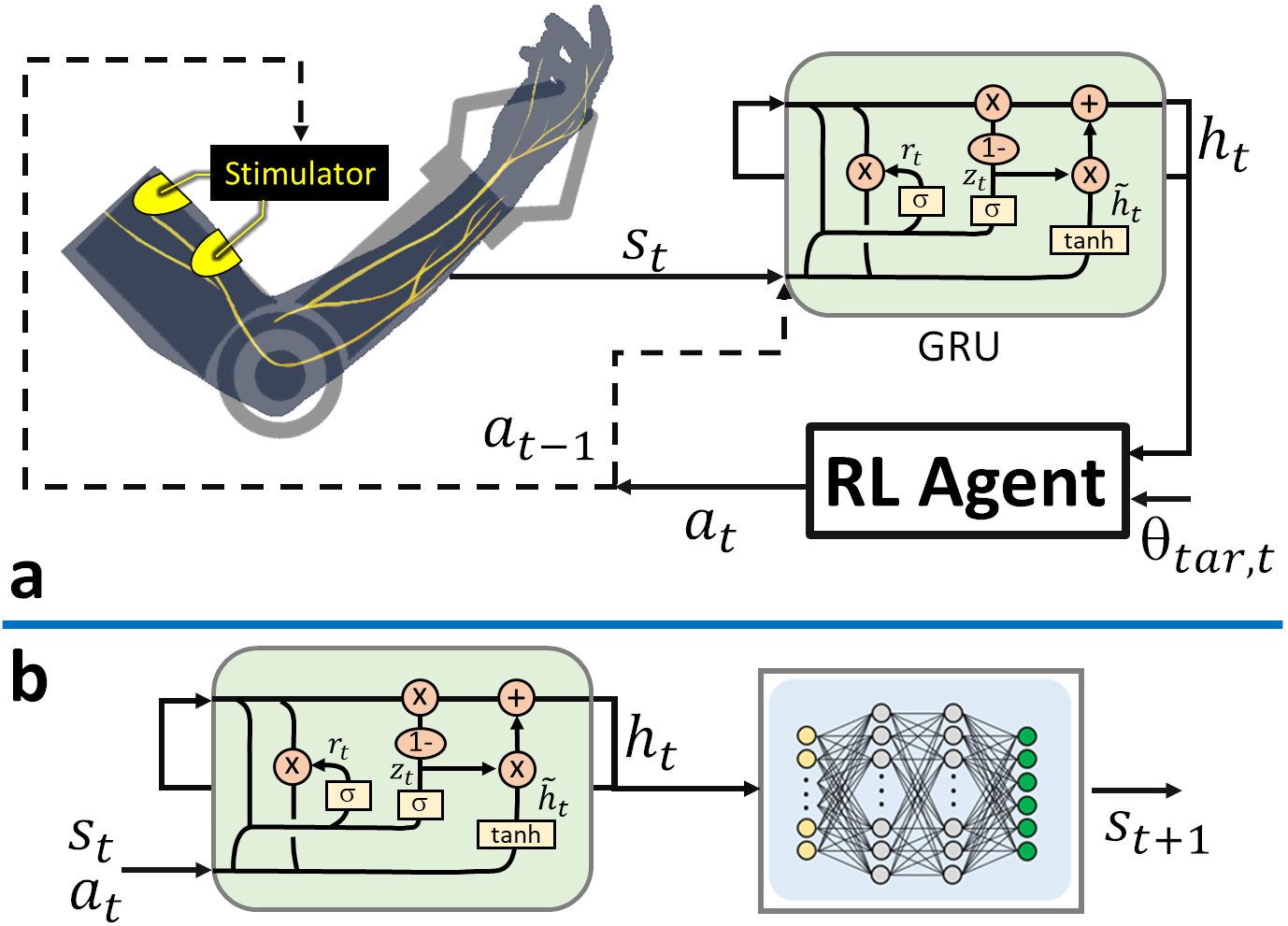}
    \caption{(a) Interactions between the components in the system during real-time, online control. Only the recurrent layer (GRU) of the state representation unit is used. (b) A fully-connected layer is added after the recurrent layer to map the hidden states to the observable states in the training of the state representation unit.}
    \label{fig:methods}

\vspace{-5pt}
\end{figure}
Controlling the movement of a human body through FES is challenging as muscle response to stimulation is non-linear and difficult to predict and model \cite{Bolsterlee2013}. Due to this, FES systems are typically controlled through simple model-free feedback controllers such as PID \cite{Dunkelberger2020}. A well-tuned PID controller can work well at the beginning, however, its performance will degrade over minutes in a single session as the response of the muscle to the stimulation is non-stationary due to short-term phenomena, muscle fatigue, but also sweating at the electrode-skin interface that affects stimulation currents, and on longer time-scales changes in muscle strength or changes in clothing e.g. for changing weather conditions such as heavy rain clothing  lighter sun clothes \cite{Thomas2008}.
To overcome this challenge to FES control, the controller should be able to learn how to elecrostimulate  so as to achieve a desired limb movement. Moreover, a learning FES system would be easier to use and could facilitate the set up at home of patients requiring only minimal or no intervention of an technically trained expert. 
This would open FES control to more advanced FES-based assistive robots, as well as the possibility of controlling human motion as we do on a robot, to restore lost motor functionality but also to augment existing ones.

Early development of FES control employed open-loop control strategies in which the stimulation is applied according to pre-defined patterns \cite{Kralj1973}.
The current development  is centred around closed-loop strategies. One challenge in designing closed-loop controllers is the optimisation of the controllers' parameters when optimal values vary across users. Several studies have, therefore, investigated closed-loop control strategies with adaptive controller parameters: Freeman \cite{Freeman2015} employed Iterative Learning Control, which is effective in producing a repetitive motion, but difficult to apply to control general open-ended motions such as reaching. Farhoud et al. \cite{Farhoud2014} and Nekoukar \cite{Nekoukar2020} used Fuzzy Logic Control to simultaneously modulate the current and pulse width of  stimulation, enabling these systems to cope with  fatigue. Yet, these systems require manually-defined active muscle patterns that specify which muscles are to be stimulated at any given state and time. This requirement limits their applications to well-studied motions such as periodic motions in cycling for which patterns are known.

Advances in machine learning in the past decade have resulted in more data-driven methods for control engineering, specifically Reinforcement Learning (RL), where an agent learns by interacting with its environment, presents great potential in control applications as substantiated by a number of successes in controlling complex real-world systems such as robots \cite{Haarnoja2018}. In FES applications, Thomas et al. \cite{Thomas2008} and Izawa et al. \cite{Izawa2004} studied the uses of RL in human arm control in simulation; Febbo et al. \cite{DiFebbo2018} used RL to control the elbow angle in human volunteers. These studies, however, investigated only short, point-to-point motions. We previously presented a  study of RL for FES cycling \cite{wannawas2021}, where an RL controller can learn to stimulate multiple muscles to track desired cycling cadences under advancing muscle fatigue. That RL system was the basis of our racing FES cycling system, with which our human paralysed athlete won the Silver medal at the last bionic Olympics, the Cybathlon 2020 held in Z\"urich, Switzerland \cite{cybathlon2020} in a 1200 m long FES bike cycle race. That controller, however, estimates the fatigue level using a hand-crafted model instead of learning it from the data. Yet, in order for RL controllers to generally work in real-world, the problem of unobservable muscle fatigue, which poses a substantial limitation on the learning and the performance of RL, needs to be addressed.

Here, we introduce a robot learning based solution that can learn to apply FES to actuate limbs, while compensating for fatigue, by using an end-to-end machine learning pipleline. For this purpose we combine Recurrent Neural Networks (RNN) and Reinforcement Learning (RL). The RNN's purpose is to use its internal states to account for the history of states and actuation controls to overcome the issue that physiologically speaking muscle state is not fully observable.
We first present our method and its formulation on example control tasks performed on neuromechanical simulations using the biomechanics simulator OpenSim that we enhanced with neurostimualtion features. We then present the performance test of our controller and the comparison with a conventional method in simulation using detailed musculoskeletal models. Finally, we present the control capability demonstrated experimentally in human volunteers by our RL-FES approach.

\vspace{-0.1cm}
\section{METHODS}
\label{sec:methods}
We first describe 3 use scenarios  and the musculoskeletal models used in the simulation. Then, we present the components of our RL controller followed by the formulation of our controller. Finally, we describe our experimental set-up for human volunteers.

\vspace{-0.1cm}
\subsection{Control Tasks and Neuromuscular Models}
\label{sec:musculoskeletal_model}
We demonstrate the use of our controller in three  FES control cases: vertical arm motion, horizontal arm motion, and cycling. The simulation study is set-up using musculoskeletal models in OpenSim \cite{Seth2011}, an open-source musculoskeletal simulation software. The models use a Hill-type muscle model based on \cite{Millard2013}. Muscle activation values are ranged between $0$ and $1$, with $1$ representing fully activated. The activation is dictated by the excitation, which is equivalent to the applied FES current intensity, and follows a first-order excitation-activation dynamics model \cite{Millard2013}. The effect of muscular fatigue is simulated using a muscular fatigue model \cite{Xia2008}. That fatigue model computes the loss of force generation capability from the history of muscle activation and fatigue rate coefficient. The three control cases have different model configurations and controlled/stimlated muscles, described as follows.
\emph{Vertical arm motion --} (Fig. \ref{fig:setup} a) The model comprises 6 muscles, 3 forming the Biceps group and 3 for the Triceps group. The musculoskeletal model has two joints, the shoulder and the elbow, all motions are within a saggital plane. The shoulder joint is locked. The elbow joint can rotate about an axis perpendicular to the vertical plane. The task here is to control the elbow angle; the controlled muscles are all in Biceps group. All muscles in Biceps group are equally excited to simulate the situation in which FES is applied via a single pair of electrodes. 
\emph{Horizontal arm motion --} (Fig. \ref{fig:setup} c) In this case motions are within a transversal plane. The arm here is supported by rollers, allowing the arm to move on a horizontal plane like moving on a table, with gravity compensation. The task is to control the elbow angle; both Biceps and Triceps groups are controlled using two separated pairs of electrodes.
\emph{Cycling motion --} (Fig. \ref{fig:setup} d) We use a lower extremity model with feet are attached to a cycling crankset that rotates without friction or ground resistance, all motions are again within two parallel saggital planes for the left and right leg respectively. The model has 18 muscles in total; 6 muscles which are rectus femoris, gluteus maximus, and hamstrings on both legs are controlled (six pairs of electrodes); the other muscles are inactive but induce resistive forces due to the elasticity of the muscles. The control task here is to control the cycling cadence.

\vspace{-0.1cm}
\subsection{Control Learning Unit}
\label{sec:control_learning_unit}
The Control Learning unit is responsible for learning a control policy, i.e., the map between the state of the system and the control signal. The control learning unit is built on the Soft Actor-Critic (SAC) framework \cite{Haarnoja2018}, a model-free RL algorithm successfully applied to solve complex control tasks in real-world robotics. This framework maximises an objective function that includes both the expected return and the entropy of the policy. The entropy term helps the learning agent balance exploration and exploitation. 

The control problem is modelled as a Markov Decision Process (MDP): At time step $t$, the agent receives a state vector $\mathbf{s}$. The agent then outputs an action vector $\mathbf{a}$, which is the stimulation intensity of the controlled muscles. The goal of the agent is to learn a policy $\pi$ that maximises an objective function computed as Equation \ref{eq:SAC_objectiveFunction}.

\vspace{-10pt}
\begin{equation}\label{eq:SAC_objectiveFunction}
\begin{gathered}
J(\pi) = \sum\limits_{t=0}^{T}\E_{\tau}[r(s_t,a_t) -
\underbrace{\alpha_{t}\log\pi_t(\boldsymbol{a_t} \vert \boldsymbol{s_t})}_\text{Entropy term}]\\
\pi^*=\arg\max_{\pi}J(\pi)
\end{gathered}
\vspace{-3pt}
\end{equation}

\vspace{-0.1cm}
\subsection{State Representation Unit}
\label{sec:state_representation_unit}
As muscle fatigue is not directly observable, in theory, we should use  partially observable MDPs (POMDP) - which is computationally demanding and data hungry \cite{li2018actor}. Therefore, we use a fully observable MDP framework and instead introduce a state representation unit which learns the hidden states of the muscles. Here, we use a recurrent neural network varient, the Gated Recurrent Unit (GRU) \cite{Chung2014}, to learn the hidden state from the time-series data. GRUs have shown consistent state-of-the-art performance in many time-series applications such as monitoring nonlinear deterioration processes \cite{Chen2019b}. The GRU combined with a fully connected output layer is trained to predict the observable state of the next time-step from the observable state and action of the previous time-step. The hidden state of the GRU is used as the state vector observed by the RL agent.

\begin{figure}[]
    \centering
    \includegraphics[width=0.98\columnwidth]{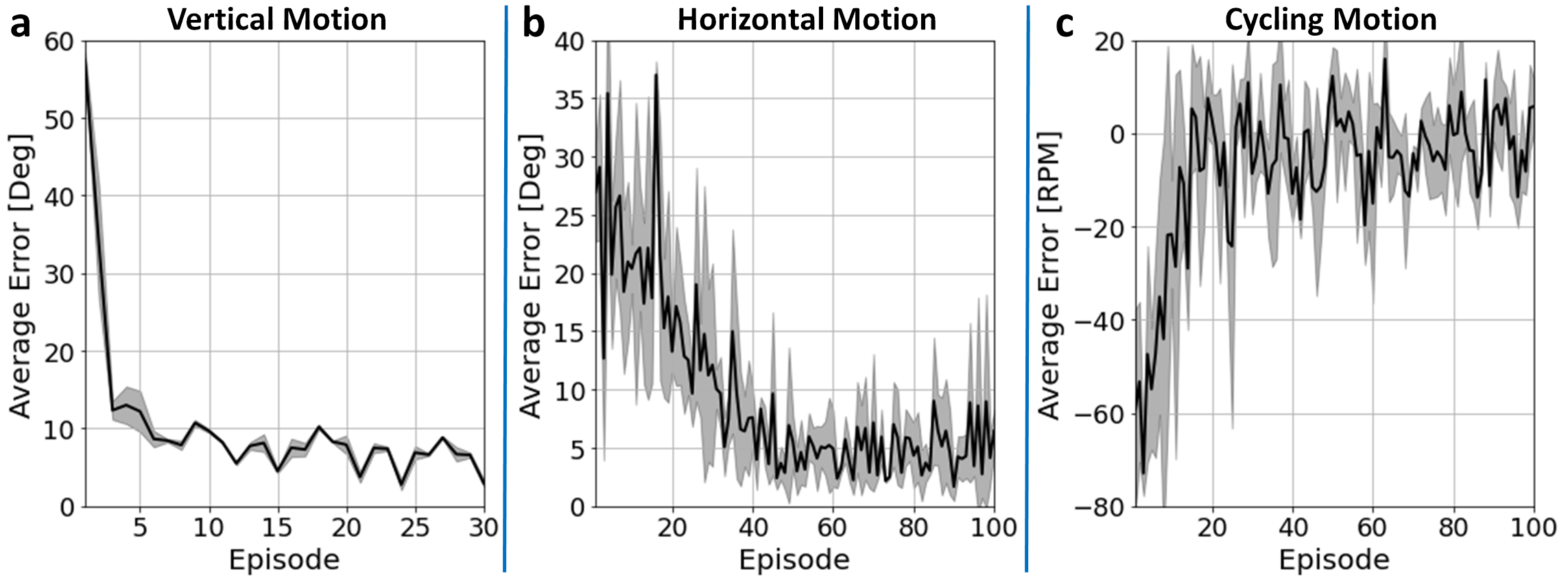}
    \caption{Average tracking error in (a,b) degrees for arm motion cases and in (c) RPM for cycling cases, against training episodes. The solid line and the shaded area show the average and the standard deviation of three training runs, respectively.} \label{fig:learning_curve}
\vspace{-5pt}
\end{figure}

\vspace{-0.1cm}
\subsection{Controller Setup and Learning}
\label{sec:controller}
The state representation and control learning units work together as a controller for which the working principle and the setup on the vertical control case are described as follows. At time-step $t$, the controller observes a state vector of the environment $\mathbf{s_t}\in\mathbb{R}^2$ which consists of elbow angle $\theta_t$ and angular velocity $\dot{\theta}_t$. The state vector is concatenated with the action $\mathbf{a_{t-1}}$ of the previous time-step and passed to the GRU which then outputs a hidden state vector $\mathbf{h_t}\in\mathbb{R}^{20}$. Note that the number of hidden states of 20 is selected empirically. We observed if this is less than 10, the controller does not learn to deal with fatigue while if it is more than 30, it slows down learning because of the high dimensional state space. The hidden state vector is concatenated with the target angle $\theta_{tar,t}$ and passed to the RL agent which outputs an action and receives a numerical reward, $r_t$. The reward function is the minus absolute difference between the actual angle and the target angle in $rad$ computed as $r(t)=-\lvert \theta_{tar,t} - \theta_t \rvert$.

The controller learns interactively from the time-series experience, wich is collected as \{$a_0,s_0,\theta_{tar,t},r_0,...,a_n,s_n,\theta_{tar,n},r_n$\}. Note, that the transition experiences are not stored in an experience replay buffer as is typically the case with standard model-free RL because, given an environment transition, the state vector observed by the RL agent changes over time as the state representation unit learns. At the end of each training episode, the learning starts with the supervised learning of the state representation unit so that it predicts $\theta_{t+1}$, and $\dot{\theta}_{t+1}$ from $\theta_{t}, \dot{\theta}_{t}, a_t$ based on the collected time-series experience. After that, the transitions observed by the RL agent, which is the hidden states of the GRU, are regenerated by feeding the sampled time-series experience to the state representation unit. The collected hidden states are then concatenated with the target angle $\theta_{tar,t}$ to form the state transitions for training the RL agent. Additionally, to accelerate the learning, we use a modification of the Hindsight Experience Replay idea \cite{Andrychowicz2017} as follows. For each series of the transitions, we create another series whose $\theta_{tar,t}$ is set to be equal to the actual angle $\theta_{t}$ throughout the series. These artificial series have the transitions with high rewards which help provide examples of good states and actions. Both real and artificial transitions series are then stored in a temporary replay buffer and the learning of the RL agent is performed following standard procedure.

The training of the controller is episodic. An episode has 1800 steps which is equivalent to 3 minutes of interaction in the real-world. The time-step size is 100 ms to cover the delayed response of the muscle. The target angle or cadence is randomly specified every 50 time-steps, which is equivalent to 5 seconds in the real-world. The controlled muscles have different initial fatigue levels and rates. The initial fatigue is level sampled from a uniform distribution between 0\%-30\%. The fatigue level of 30\% means the muscle can produce 70\% of the maximum force when stimulated with maximum intensity. The fatigue rate is also sampled from a uniform distribution with the upper and lower bounds which result in the level of fatigue reaching 50\% when it is stimulated with maximum intensity for 1 and 2 minutes, respectively. These rates are approximated from \cite{Kralj1973} and rechecked with our empirical experiments. Throughout an episode, the fatigue level changes as a function of applied stimulation intensity and the fatigue rate. As a comparison baseline, we also implement a PID controller on all our tasks, due to it being the conventional method used in many FES control applications such as cycling \cite{DeSousa2016, Woods2018} and arm motion \cite{Wolf2020}. In tasks requiring control of multiple muscles, such as cycling, PID control is used together with a pre-defined active muscle pattern that determines to which muscles the stimulation is applied.

\vspace{-0.1cm}
\subsection{Human Volunteer Experimental Setup}
To validate our results we investigated the capability of our framework using a set-up that implements the arm vertical control as described in Sec.~\ref{sec:musculoskeletal_model}. We used a Hasomed Rehastim 1 (HASOMED GmbH, Magdeburg, Germany) stimular. The stimulator is computer controlled via an API and generates biphasic FES pulses with a fixed frequency of 35 Hz. Our control signal  is the stimulation intensity, which can be manipulated both in terms of pulse amplitude and pulse width. Attempting to control both of these complicates the control problem as e.g. varying only the current level will  lead to sharp, non-monotonous responses, while varying pulse only will quickly result in fatigue and discomfort. To remedy this and enhance safety, we use a set of 20 consecutively ordered  stimulation intensities that are made up of safe tuples of pulse current and width. We stimulate the Biceps muscles via pairs of 4x6 cm oval surface electrodes. The elbow angle is tracked through an IMU (LSM9DS1) embeded with an Arduino Nano Sense which gives real-time feedback. The RL controller was trained in simulation for 30 episodes before transferred into real-world use. The gains of the PID controller are manually fine-tuned in the real-world setup to optimise performance. Both controllers are tested on two three-minute-long tracking tasks similar to those in simulation. The results were collected on different days to minimise the effect of muscle fatigue.

\vspace{-0.1cm}
\section{RESULTS \& DISCUSSION}
\label{sec:results}
We verify the use of our controller in simulation, considering three scenarios: 1. Vertical arm motion, 2. Horiztonal arm motion and 3. Cycling motion. As a proof of concept, we also test scenario 1 in the real-world, with one participant.
\subsection{Simulation Results}
\vspace{-0.2cm}
\label{sec:sim_results}
\emph{Arm vertical motion -- } The controller is trained for 30 episodes to track random angle trajectories. (Fig. \ref{fig:learning_curve} a) The learning of the controller is monitored by average tracking error over an episode. The plot shows that the controller can produce low average error within 10 episodes. Yet, the controller is still unable to deal with muscle fatigue at this stage for it produces positive error in the first and negative error in the second half of an episode. It takes around 25 episodes for the controller to consistently produce low error throughout an episode.

\begin{figure*}[]
    \center
    \includegraphics[width=0.95\textwidth]{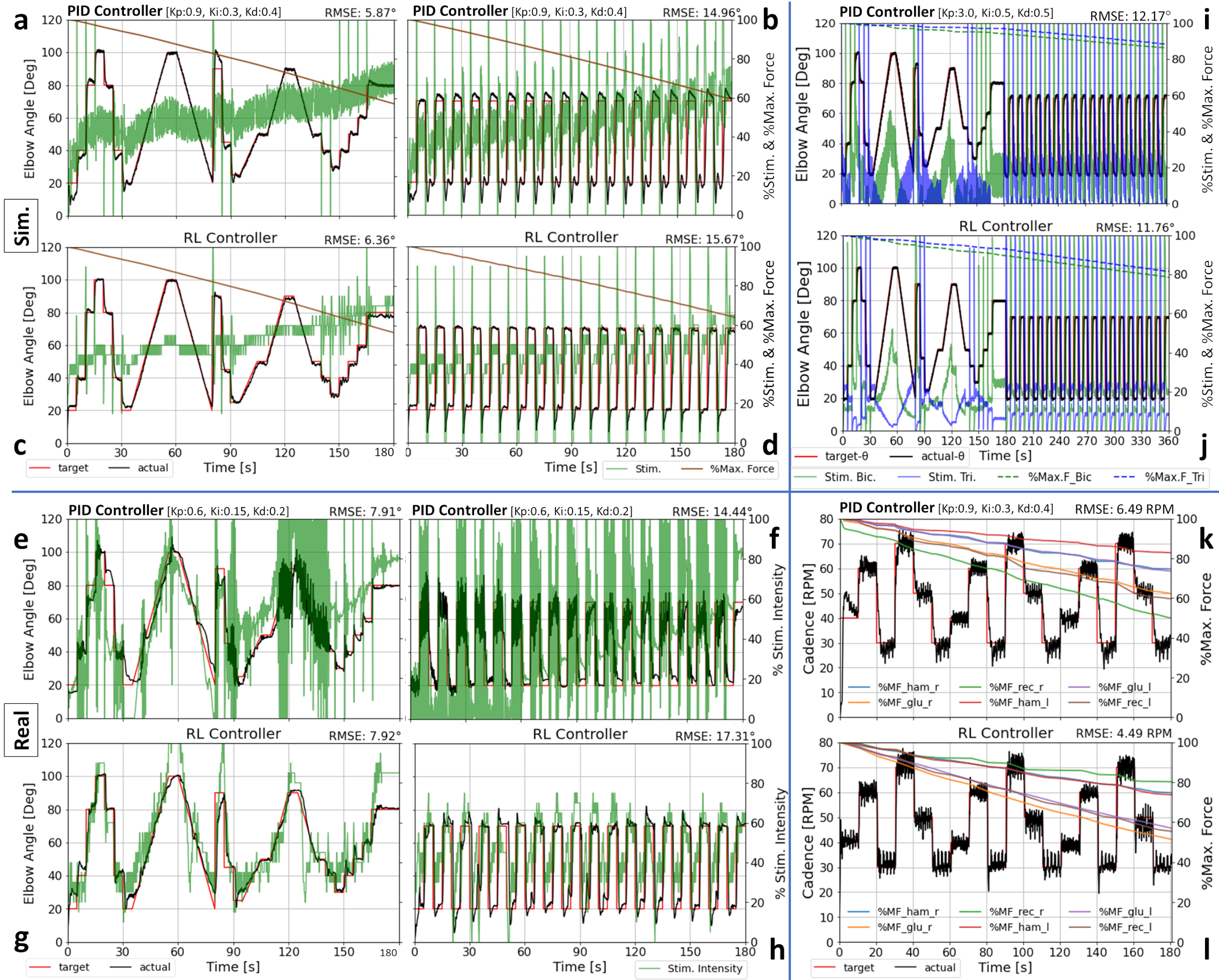}
    \caption{\label{fig:Results}Simulation results of vertical arm motion task of (a,b) PID and (c,d) RL controllers. Experimental results of (e,f) PID and (g,h) RL  in the same task. Simulation results of horizontal arm motion of (i) PID and (j) RL controllers. Results of cycling motion of (k) PID and (l) RL controllers. \%Max. force in (a-h), shorten to \%MF in (i-l), show the decline of maximum muscle force due to fatigue.}
    \vspace{-0.3cm}
\end{figure*}
We compare  RL and PID controller performance  in 2 three-minute-long tracking tasks. The first task is to track arbitrary angle trajectories that include ramping. Fig.~\ref{fig:Results}.a,c show the performance of  PID and RL controllers, respectively. Both controllers track the trajectory. The PID controller has insignificantly better performance in root mean square error (RMSE). However, the RL controller produces fewer overshoots and  better prepares for the drop of the arm as highlighted by blue circles in Fig. \ref{fig:Results} a,c. The second task is to track a trajectory that periodically sets the angle at 70 and 20 degrees. These two angles are selected to be in regions that are relatively difficult to maintain. Again, both controllers are able to track the trajectory in this task, as seen in Fig. \ref{fig:Results} b and d. The overshoots of the PID controller increase as the muscle becomes more fatigued, evidenced by the downward trajectory of the brown lines which show the percentage maximum force that the muscles can produce -- an established fatigue effect \cite{Xia2008}. Note that  overshoots grow because the angle increases slower than expected when the muscles become fatigue, thereby causing the accumulated error that results in a large control signal. Therefore, the PID tuning that produces good control in earlier periods cannot maintain the performance in a later one. The RL controller has a more consistent performance in term of the overshoots. Its ability to prevent the drop of the angle below $20^{\circ}$ decreases as the muscles become more fatigued. However, here it still performs better than the PID controller.

\emph{Arm horizontal motion --} The task  is to control elbow angle of the arm moving on the horizontal plane by stimulating Biceps and Triceps muscles. Fig.~\ref{fig:learning_curve} b shows the plot of average error against training episode. The controller takes around 50 episodes to converge and is then tested to track a 6-minute long trajectory (Fig.~\ref{fig:Results} j). The RL controller is able to track the trajectory with slightly better average error than the PID controller (Fig.~\ref{fig:Results} i) which has growing overshoots as the muscles become fatigued. Additionally, the RL controller learns to apply co-contraction to prevent the overshoot and keep the angle still at the target.

\emph{Cycling --} The task in this case is to control cycling cadence by stimulating 6 muscles: rectus femoris, gluteus maximus, and hamstrings on both legs. The plot of the average error against training episodes (Fig. \ref{fig:learning_curve} c) shows that the controller takes around 30 episodes to produce low error. Yet, similar to the vertical motion case, it takes around 60 episodes to learn to deal with the fatigue. The average error of RL controller (\ref{fig:Results} l) is slightly better than that of the PID controller (\ref{fig:Results} k) with the active muscle pattern adopted from \cite{DeSousa2016}. RL controller, on the other hand, learns both stimulation intensity and the active-muscle pattern as a whole on its own.

\vspace{-0.1cm}
\subsection{Human volunteer experimental results}
\vspace{-0.2cm}
We  validate our RL approach in real-world human volunteer experiment . The RL controller was trained for 30 episodes in simulation before being tested in the real-world. The RL controller is then tested on the vertical arm motion, following the same tracking requirements as described in the simulation results section. Both controllers perform poorer in real-world than in simulation. In the first desired trajectory the PID controller (Fig.~\ref{fig:Results} e) produces good tracking in the first half but unstable control in the second half of the trajectory, especially along the ramping section. Our RL controller produces more consistent control along the same trajectory (Fig.~\ref{fig:Results} g). In the second case, the PID controller (Fig. \ref{fig:Results} f) produces unstable control at the 70$^{\circ}$ target. However, it still has a slightly better RMSE. The RL controller, again, has more consistent control (Fig.A\ref{fig:Results} h).

\section{CONCLUSIONS} 
\vspace{-0.1cm}
\label{sec:conclusions}
We present a controller that can learn to control the non-linear and non-stationary actuators that are human muscles, through functional electrical stimulation (FES). This effectively allows us to control human motion, as we control robot motion, with applications in assistive and rehabilitation robotics \cite{Shafti2021, Maimon-Dror2017}, as well as robotic human augmentation. Our controller uses a combination of reinforcement learning (RL) and recurrent neural networks, specifically gated recurrent units (GRU), to learn non-stationary hidden states of human muscles. We then formulate a fully-observable RL problem by letting the RL agent perceive the state of the environment through the hidden states of the GRU. 

We demonstrate the uses of our controller in controlling the movement of human limbs by inducing muscle contraction using Functional Electrical Stimulation (FES). We first work in simulation using detailed musculoskeletal models in various tasks: vertical and horizontal motions of the arm, as well as cycling motion. We test the performance of our controller in tracking different trajectories in these tasks. We compare the performance of our controller to that of a PID controller, a conventional method for controlling FES \cite{Wolf2020, Lynch2008a}. In single muscle control scenario of vertical motion, the simulation results show that both controllers can track desired trajectories. The PID and RL controller have matching RMSE tracking error, however, the RL controller shows far less overshoots and more consistent performance under fatigue. Our RL controller starts to outperform the PID controller when the task involves the control of multiple muscles as illustrated by the horizontal arm motion and the cycling scenarios. This substantiates the further potential of RL controllers in more complex cases such as the full control of arm motion.

We also evaluate the performance of our controller in human volunteers confirming our simulation results for vertical arm control.  We evaluate the performance of our controller and PID controller in the same tracking tasks. The performance of both controllers is in human volunteers, not surprisingly for real-world deployments poorer than in simulation. However, the PID controller has unstable control at some parts along the trajectory, while our RL controller shows more stability and consistent performance throughout the repeated trajectories as muscles fatigue. Our state representation unit combination with Deep RL can of course be used also for other non-stationary, non-linear actuators such as pneumatic muscles and soft robotic actuators.

In summary, our FES-RL results open the way for thinking about assistive robotics in new ways. This is important as FES in paralysed users has already shown to help build muscle, improve range of limb motion and cardiovascular health, and gain in life expectancy \cite{nightingale2007benefits}. Crucially, by restoring function to one's own actuators, the end-users sense of agency and thus embodiment and adoption of technology may be greatly increased \cite{makin2017neurocognitive}.





\vspace{-0.1cm}
\section*{ACKNOWLEDGMENTS}
Funding by a Royal Thai Govt. Scholarship to NW and a UKRI Turing AI Fellowship (EP/V025449/1) to AAF. 


\vspace{-0.1cm}
\bibliographystyle{ieeetr}
\bibliography{IROS21Ref}

\end{document}